\documentclass[runningheads]{llncs}

\usepackage{eccv}

\usepackage{eccvabbrv}

\usepackage{graphicx}
\usepackage{booktabs}
\usepackage{xcolor}
\usepackage{multirow}
\usepackage{amsmath}

\usepackage[breaklinks,colorlinks]{hyperref}

\usepackage[accsupp]{axessibility}

\newcommand{\methodname}{ISO}

\usepackage{hyperref}

\usepackage{orcidlink}

\begin{document}

\title{Monocular Occupancy Prediction for Scalable Indoor Scenes} 

\author{Hongxiao Yu\inst{1,2}\orcidlink{0009-0003-9249-2726} \and
Yuqi Wang\inst{1,2}\orcidlink{0000-0002-6360-1431} \and
Yuntao Chen\inst{3}\orcidlink{0000-0002-9555-1897} \and
Zhaoxiang Zhang\inst{1,2,3}\orcidlink{0000-0003-2648-3875}}

\authorrunning{H.~Yu et al.}

\institute{
School of Artificial Intelligence, University of Chinese Academy of Sciences (UCAS) \and
NLPR, MAIS, Institute of Automation, Chinese Academy of Sciences (CASIA) \and
Centre for Artificial Intelligence and Robotics (HKISI\_CAS) \\
\email{\{yuhongxiao2023, wangyuqi2020, zhaoxiang.zhang\}@ia.ac.cn, chenyuntao08@gmail.com}
}

\maketitle

\begin{abstract}
    Camera-based 3D occupancy prediction has recently garnered increasing attention in outdoor driving scenes. However, research in indoor scenes remains relatively unexplored. The core differences in indoor scenes lie in the complexity of scene scale and the variance in object size.
    In this paper, we propose a novel method, named \methodname{}, for predicting indoor scene occupancy using monocular images. \methodname{} harnesses the advantages of a pretrained depth model to achieve accurate depth predictions. Furthermore, we introduce the Dual Feature Line of Sight Projection (D-FLoSP) module within \methodname{}, which enhances the learning of 3D voxel features.
    To foster further research in this domain, we introduce \emph{Occ-ScanNet}, a large-scale occupancy benchmark for indoor scenes. With a dataset size 40 times larger than the NYUv2 dataset, it facilitates future scalable research in indoor scene analysis.
    Experimental results on both NYUv2 and Occ-ScanNet demonstrate that our method achieves state-of-the-art performance. 
    The dataset and code are made publicly at \url{https://github.com/hongxiaoy/ISO.git}.

    \keywords{3D occupancy prediction \and Camera-based scene understanding \and Semantic scene completion}
\end{abstract}

\section{Introduction}
\label{sec:intro}

3D scene understanding is a crucial task in computer vision, becoming increasingly important for applications such as robotic navigation~\cite{desouza2002vision}, augmented reality~\cite{azuma1997survey}, and autonomous driving~\cite{yurtsever2020survey}. While humans possess a natural ability to comprehend 3D environments through vision, enabling this ability in computers poses a significant challenge due to limitations like restricted fields of view, sparse sensing capabilities, and measurement noise. However, with the rapid development of deep learning and the availability of large-scale 3D driving datasets, camera-based 3D object detection~\cite{wang2022detr3d, li2022bevformer, huang2021bevdet, li2023bevdepth, wang2023frustumformer, yang2023bevformer, wang2023exploring, guan2024gramo} and occupancy prediction~\cite{cao2022monoscene, tian2024occ3d, li2023fb, wei2023surroundocc, wang2024panoocc} have developed rapidly, achieving significant improvements in performance.

Occupancy prediction has gained popularity recently due to its effectiveness in representing both background and foreground objects within a scene using a unified representation. Although significant progress has been made in outdoor driving scenarios, research in indoor scenarios remains limited. Indoor scenes differ from outdoor driving scenes in two key aspects:
(1) \emph{scene-scale complexity}: Indoor rooms often exhibit a more diverse range of sizes compared to outdoor environments, where driving scenarios typically focus on a fixed 3D space for perception. This diversity, ranging from spacious living rooms to narrow kitchens, poses higher precision requirements for depth prediction.
(2) \emph{object complexity}: Indoor scenes feature a higher density and greater variety of objects. Unlike outdoor objects such as vehicles and pedestrians, which typically have consistent sizes within their respective categories and are relatively well-separated, indoor furniture and other objects often exhibit significant variations in scale and are closely positioned to each other. This increased complexity necessitates more sophisticated 3D perception techniques to accurately capture and understand the intricate geometry and relationships among objects within indoor spaces. Furthermore, existing works \cite{cao2022monoscene, yao2023ndc} focusing on indoor scenes primarily utilize the NYUv2 dataset \cite{silberman2012indoor}, which lacks a more scalable and generalizable benchmark for comprehensive evaluation.

To address the above issues, we firstly introduce a more scalable benchmark, \emph{Occ-ScanNet}, for 3D occupancy prediction in indoor scenes. This benchmark builds upon the large-scale ScanNet~\cite{dai2017scannet} dataset, offering a 40 times more samples compared to the NYUv2~\cite{silberman2012indoor} dataset, thus greatly expanding the scope and diversity of indoor scene study. 
To effectively address the unique complexities of indoor scenes, we propose a novel method named \methodname{} (Indoor Scene Occupancy). \methodname{} leverages a powerful pretrained depth model and integrates a D-FLoSP (Dual Feature Line of Sight Projection) module. This module enables precise depth estimation and facilitates learning voxel features for accurate predictions.
To accommodate the varying sizes of indoor objects, we introduce a multi-scale feature fusion module. This module enhances the learning of object features across different scales.
Leveraging a powerful depth model, our method is adept at handling diverse indoor scenes, offering a robust and versatile solution.

Our main contributions can be summarized as follows.
\begin{itemize}
    \item We introduce a new benchmark called \emph{Occ-ScanNet} for monocular 3D occupancy prediction in indoor scenes. With a dataset size 40 times larger than the NYUv2 dataset, it significantly enhances the potential for scalable research in indoor scene analysis.
    \item We propose a novel approach called \emph{ISO}, which primarily comprises the D-FLoSP (Dual Feature Line of Sight Projection) module and a multi-scale feature fusion module. Together, these components effectively address the challenges posed by variations in scene and object sizes, enabling more accurate and robust 3D occupancy prediction.
    \item Experiments on the Occ-ScanNet and NYUv2 datasets demonstrate that our approach achieves state-of-the-art performance in monocular 3D occupancy prediction. Furthermore, our method exhibits significant scalability potential.
\end{itemize}

\section{Related works}
\subsection{Monocular 3D Semantic Scene Completion}
Monocular 3D Semantic Scene Completion (SSC) aims to infer the complete 3D structure and corresponding semantics from a single image. This monocular setting was first introduced by Monoscene~\cite{cao2022monoscene}, advancing upon prior SSC methods~\cite{song2017semantic, zhang2018efficient, liu2018see, li2020anisotropic, zhong2020semantic} by relying solely on vision, without additional 3D inputs.
Monoscene~\cite{cao2022monoscene} introduces the Features Line of Sight Projection (FLoSP), inspired by optics, to obtain 3D features through ray projection. However, the shared 2D features lifted to 3D rays via FLoSP results in depth ambiguity. Consequently, subsequent efforts have placed greater emphasis on leveraging depth information. 
VoxFormer~\cite{li2023voxformer} proposed a novel query proposal network based on 2D convolutions, generating sparse queries from image depth, which showed impressive performance in driving scenes.
Meanwhile, NDC-Scene~\cite{yao2023ndc} further devised a Depth-Adaptive Dual Decoder to concurrently upsample and merge the 2D and 3D feature maps, thereby enhancing overall performance. 

\subsection{Multiview 3D Occupancy Prediction}

Multiview 3D Occupancy Prediction predicts the semantic occupancy of the surrounding 3D scene given multiview images, has recently garnered significant attention in the field of autonomous driving. TPVFormer~\cite{huang2023tri} pioneers exploration in driving scenes, employing sparse LiDAR labels for supervision. It introduces a tri-perspective view (TPV) representation that accompanies BEV with two additional perpendicular planes. However, due to the sparse supervision provided by LiDAR, subsequent works~\cite{tian2024occ3d, wang2023openoccupancy,tong2023scene,wei2023surroundocc} have focused on providing more dense occupancy benchmarks. FB-OCC~\cite{li2023fb} integrates the lifting of 2D to 3D and querying of 3D to 2D, achieving a more efficient 3D feature transformation. Meanwhile, PanoOcc~\cite{wang2024panoocc} unifies detection and semantic occupancy tasks, enabling comprehensive panoramic scene understanding.
Recent methods also explore improving model efficiency~\cite{yu2023flashocc} and utilizing weaker forms of supervision~\cite{pan2023renderocc, huang2024selfocc}.

\subsection{3D Reconstruction from Image}
3D reconstruction is a technique that involves recovering a 3D representation of objects \cite{fan2017point,park2019deepsdf,arshad2023list}, scenes \cite{wu2023objectsdf++,zhang2022nerfusion,denninger20203d}, or even human bodies \cite{fieraru2023reconstructing,goel2023humans,zheng2019deephuman} from camera images. The task of 3D reconstruction can be classified into two categories: monocular reconstruction and multi-view reconstruction, depending on the number of images utilized. In the context of indoor scenes, 3D scene reconstruction aims to determine the surface geometry of the entire scene, often without incorporating semantic information. Atlas \cite{murez2020atlas} proposes an end-to-end 3D reconstruction framework using TSDF regression from RGB images, bypassing traditional depth map estimation for efficient semantic segmentation.
Panoptic-Reconstruction \cite{dahnert2021panoptic} unifies geometric reconstruction, 3D semantic, and instance segmentation tasks. It predicts complete geometric reconstruction, semantic, and instance segmentations from a single RGB image's camera frustum. SCFusion \cite{wu2020scfusion}, a real-time scene reconstruction framework, integrates continuous depth data using neural architecture for occupancy maps. It efficiently combines semantic completion with voxel states for simultaneous scene reconstruction and semantic understanding in real-time 3D environments.

\section{Method}

\subsubsection{Problem definition.}
We focus on the problem of monocular 3D Occupancy Prediction. Specifically, this task takes a single RGB image $\mathbf{I}^{RGB}$ as input and output a voxel-wise occupancy along with semantic categories $\mathbf{Y}^{X\times Y\times Z \times C}$. $X$, $Y$, and $Z$ represent the dimensions of the predicted 3D scene, while $C$ denotes the total number of semantic categories.

\begin{figure}[tb]
  \centering
  \includegraphics[width=\textwidth]{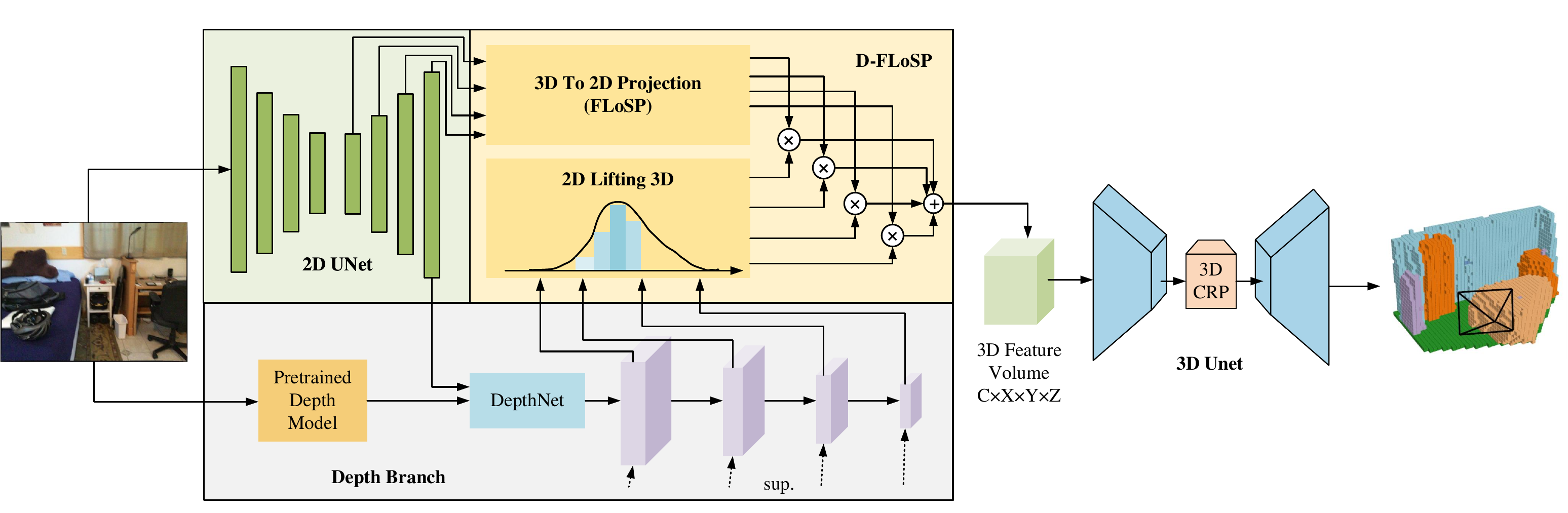}
  \caption{The core design of \textbf{\methodname{}} centers around the transformation of features from 2D to 3D spaces, encompassing the \emph{Depth Branch} and the \emph{D-FLoSP} module. A depth branch is initially integrated, it leverages a pre-trained depth model to estimate a pixel-wise depth map which is processed by the DepthNet to generate the final depth distribution. An element-wise multiplication between the voxel depth and features followed by summation are subsequently performed to derive the initial 3D voxel feature. The 3D feature is further processed to predict the 3D scene occupancy.}
  \label{fig:model_overview}
\end{figure}

\subsubsection{Model overview.}
In this section, we introduce the overall architecture of \methodname{}, as shown in~\cref{fig:model_overview}. Following Monoscene~\cite{cao2022monoscene}, we utilize 2D Unet and 3D Unet architectures to handle 2D and 3D features.
Our core design focuses on transforming features from 2D to 3D, incorporating the \emph{Depth Branch} and \emph{D-FLoSP} module.
Specifically, we incorporate a depth branch to estimate the pixel-wise depth map using a pre-trained depth model. Processed by the DepthNet, the model outputs a refined depth distribution. Then, voxel depth and features are multiplied element-wise and summed to obtain the initial 3D voxel feature volume $\mathbf{X^{3d}}\in \mathbb{R}^{X\times Y \times Z \times C}$. Finally, after processing the 3D voxel features, the model outputs the 3D scene occupancy.

\subsection{Depth Branch}
In this section, we introduce how to effectively estimate depth information from a single image.

\subsubsection{Coarse depth estimation.}
Learning depth from scratch can be quite challenging. However, thanks to the recent rapid advancements in depth estimation~\cite{ranftl2020towards, birkl2023midas, bhat2023zoedepth, yang2024depth}, we can leverage pre-trained depth models to initially estimate a coarse depth map.

Compared to past models that could only estimate relative depth, we opted for Depth-Anything~\cite{yang2024depth} as the pre-trained depth model, because it excels in predicting metric depth, as the~\cref{eq:metric_depth} shows:
\begin{align}
    \label{eq:metric_depth}
    \mathbf{D}^{\text{metric}} = \mathbf{N}_{\text{depth}}(\mathbf{I}^{\text{rgb}}) \in \mathbb{R}^{1\times H\times W},
\end{align}
where $\mathbf{N}_{\text{depth}}$ denotes the pre-trained depth model~\cite{bhat2023zoedepth,yang2024depth}, and $\mathbf{I}^{\text{rgb}}$ represents the input image. $H$ and $W$ represents the height and width of the input image.

\subsubsection{Depth refinement.}

The coarse depth estimation is subsequently refined through model learning. The depth from pre-trained depth model is not precise enough for a higher mIoU, so we design a fine-tuning strategy. Specifically, the predicted metric depth $\mathbf{D}^{\text{metric}}$ is then concatenated with the image feature $\mathbf{X}$ that has the same spatial scale. The augmented feature is continue processed by a following DepthNet to get a refined depth distribution $\mathbf{D}^{\text{dist}}_{\text{s=1}}$, as \cref{eq:depth_refine} shows:

\begin{align}
    \label{eq:depth_refine}
    \mathbf{D}^{\text{dist}}_{\text{s=1}} = \mathbf{F}_{\text{depth}}\bigg(\text{Concat}\big(\mathbf{D}^{\text{metric}},~\mathbf{X}^{\text{2d}}_{\text{s=1}}\big)\bigg),~ \mathbf{D}^{\text{dist}}_{\text{s=1}} \in\mathbb{R}^{N_{\text{bins}}\times H\times W},
\end{align}
where $\mathbf{F}_{\text{depth}}$ is the DepthNet, $\mathbf{X}^{\text{2d}}_{\text{s=1}}$ is the 2D image feature, and $N_{\text{bins}}$ is the number of discrete depth bins. $s$ denotes the level of feature scale, where $s=1$ implies that the feature is at a 1:1 original image ratio. 

We down-samples the scale $\frac{1}{1}$ distribution to get other smaller-scale 2D image feature maps' depth distribution. We represent the procedure as \cref{eq:down_sample}:
\begin{align}
    \label{eq:down_sample}
    \mathbf{D}^{\text{dist}}_{\text{s}=k} = \text{DownSample}(\mathbf{D}^{\text{dist}}_{\text{s=1}}),~k={2,~4,~8}.
\end{align}

The refined depth is supervised by the ground truth $\mathbf{D}^{\text{GT}}$. The ground truth has single depth value at each pixel, so we convert the single value to a one-hot vector of length $N_{\text{bins}}$. The refined depth prediction can be optimized using the BCE loss function as \cref{eq:depth_loss} shows.

\begin{multline}
    \label{eq:depth_loss}
    \mathcal{L}_\text{depth} = -\frac{1}{N_{\text{bins}} \times H \times W} \\ \sum_{d=1}^{N_{\text{bins}}} \sum_{h=1}^{H} \sum_{w=1}^{W} [\mathbf{D}^{\text{GT}}_{N_\text{bins},d,h,w} \cdot \log(\mathbf{D}^{\text{dist}}_{N_\text{bins},d,h,w}) \\
    + (1 - \mathbf{D}^{\text{GT}}_{N_\text{bins},d,h,w}) \cdot \log(1 - \mathbf{D}^{\text{dist}}_{N_\text{bins},d,h,w}) ]
\end{multline}

\subsection{Dual Feature Line of Sight Projection (D-FLoSP)}

Compared to the FLoSP module in MonoScene~\cite{cao2022monoscene}, which projects 3D voxels to 2D, and 3D feature vectors are the corresponding pixel feature vectors. It only considers the projection of 3D rays and overlooks depth information, our proposed D-FLoSP module can more effectively integrate depth information. 

Specifically, after getting depth distribution in the camera frame, we implement the FLoSP module to depth distribution. Each 3D voxel centroid position $x^c$ in the world frame can be projected to the camera frame and the pixel frame using camera pose and intrinsic matrix. We assign the projected depth values to discrete depth bins index by \cref{eq:range_bins,eq:depth_index}, according to \cite{tang2020center3d}, 
\begin{align}
    \label{eq:range_bins}
    \delta & = \frac{2(d_{\text{max}} - d_{\text{min}})}{N_{bins}(1+N_{bins})}, \\
    \label{eq:depth_index}
    l & = -0.5 + 0.5\sqrt{1+\frac{8(d-d_{\text{min}})}{\delta}},
\end{align}
and the depth distribution probability it lies in serves as the depth probability of that voxel. The process is illustrated by \cref{eq:3d_to_2d_sampling_depth}:
\begin{align}
    \label{eq:3d_to_2d_sampling_depth}
    \mathbf{D}^{\text{3d}}_{\text{s}=k} = \mathbf{\Phi}^{\text{3d}}_{\rho(x^c)}(\mathbf{D}^{\text{dist}}_{\text{s}=k}) \in \mathbb{R}^{X\times Y\times Z\times C}, k={1,2,4,8},
\end{align}
where $\mathbf{\Phi}_a^{\text{3d}}(b)$ is the 3D sampling of b at coordinates a, and $\rho(\cdot)$ is the perspective projection. Thus, the depth distribution is lifted from 2D to 3D. The bin-based representation with quantization is used for mapping continuous depth value to a discrete form, each bin will be assigned a different depth distribution score to make voxel aware of depth. The original depth precision is a continuous value, and hard to integrate it with the discrete form of voxel.

Following~\cite{cao2022monoscene}, we also employ the FLoSP module to generate a 3D voxel feature from a 2D feature map. Similar to what we did with the voxel centroids, the projected voxel can sample corresponding 1:k scaled feature maps $F_{1:k}$ from the 2D UNet decoder. This process is illustrated by Equation \cref{eq:3d_to_2d_sampling}:
\begin{align}
    \label{eq:3d_to_2d_sampling}
    \mathbf{X}^{\text{3d}}_{\text{s}=k} = \mathbf{\Phi}^{\text{2d}}_{\rho(x^c)}(\mathbf{X}^{\text{2d}}_{\text{s}=k}) \in\mathbb{R}^{X\times Y\times Z\times C}, k={1,2,4,8},
\end{align}
where $\mathbf{\Phi}_a^{\text{2d}}(b)$ is the 2D sampling of b at coordinates a, and $\rho(\cdot)$ is the perspective projection.

The final 3D feature map $\mathbf{X}^{\text{3d}}$ are summed in \cref{eq:3dsum}:
\begin{align}
    \label{eq:3dsum}
    \mathbf{X}^{\text{3d}} = \sum_{\text{s}\in{\{1,2,4,8\}}}\mathbf{X}^{\text{3d}}_{\text{s}=k}\odot \mathbf{D}^{\text{3d}}_{\text{s}=k},
\end{align}
where $\odot$ is the element-wise multiplication.
The output map $\mathbf{X}^{\text{3d}}$ is then serves as the input of 3D UNet.

\begin{figure}[tb]
  \centering
  \begin{subfigure}{0.49\linewidth}
    \includegraphics[width=\textwidth]{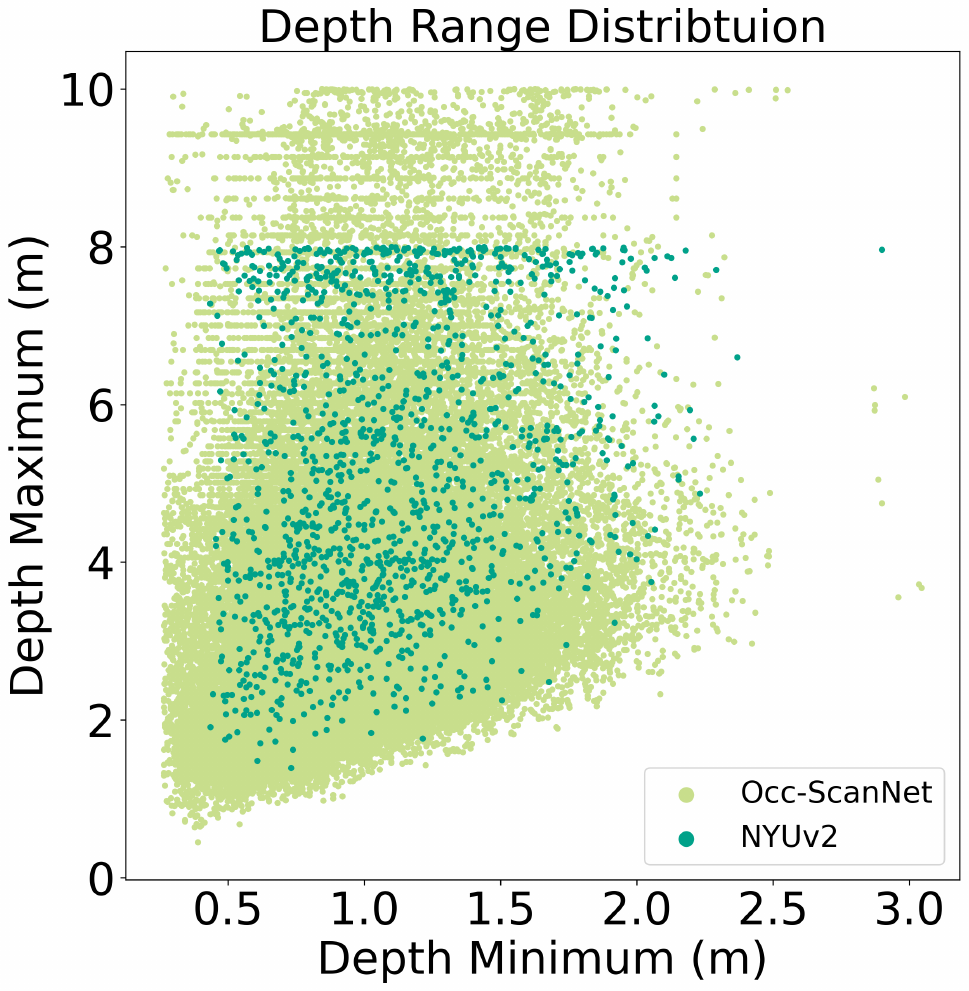}
    \caption{Depth ranges in NYUv2 and our Occ-ScanNet Benchmark}
    \label{fig:dataset_depth_distribution}
  \end{subfigure}
  \hfill
  \begin{subfigure}{0.49\linewidth}
    \includegraphics[width=\textwidth]{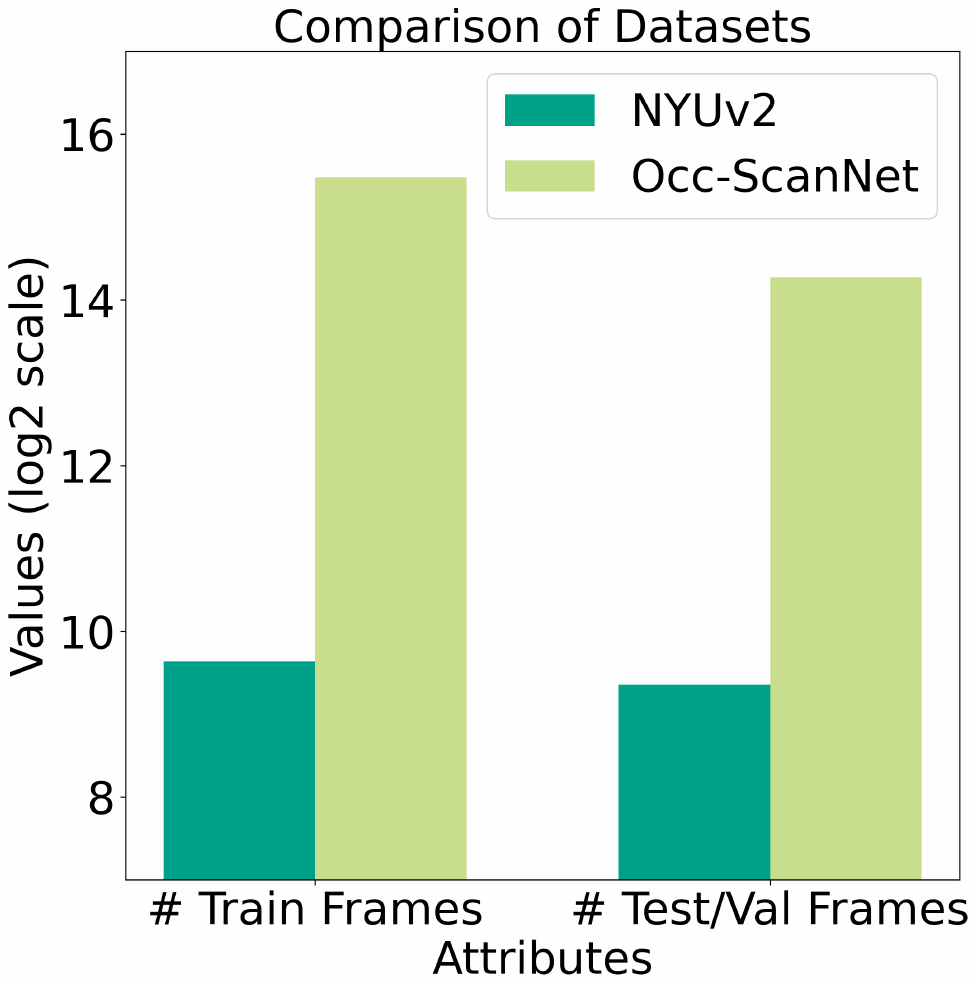}
    \caption{Dataset attributes of NYUv2 and our Occ-ScanNet Benchmark}
    \label{fig:dataset_attributes}
  \end{subfigure}
  \caption{\textbf{Comparison of NYUv2 and Occ-ScanNet Benchmark}. In (a), the depth ranges of NYUv2 and Occ-ScanNet are distinguished by dark and light green, respectively, with the horizontal axis indicating the minimum depth and the vertical axis showing the maximum depth of scenes. (b) quantitatively demonstrates that Occ-ScanNet possesses a significantly larger data scale compared to the NYUv2 dataset.}
  \label{fig:dataset_comparison}
\end{figure}

\begin{figure}[tb]
  \centering
  \includegraphics[width=1.0\textwidth]{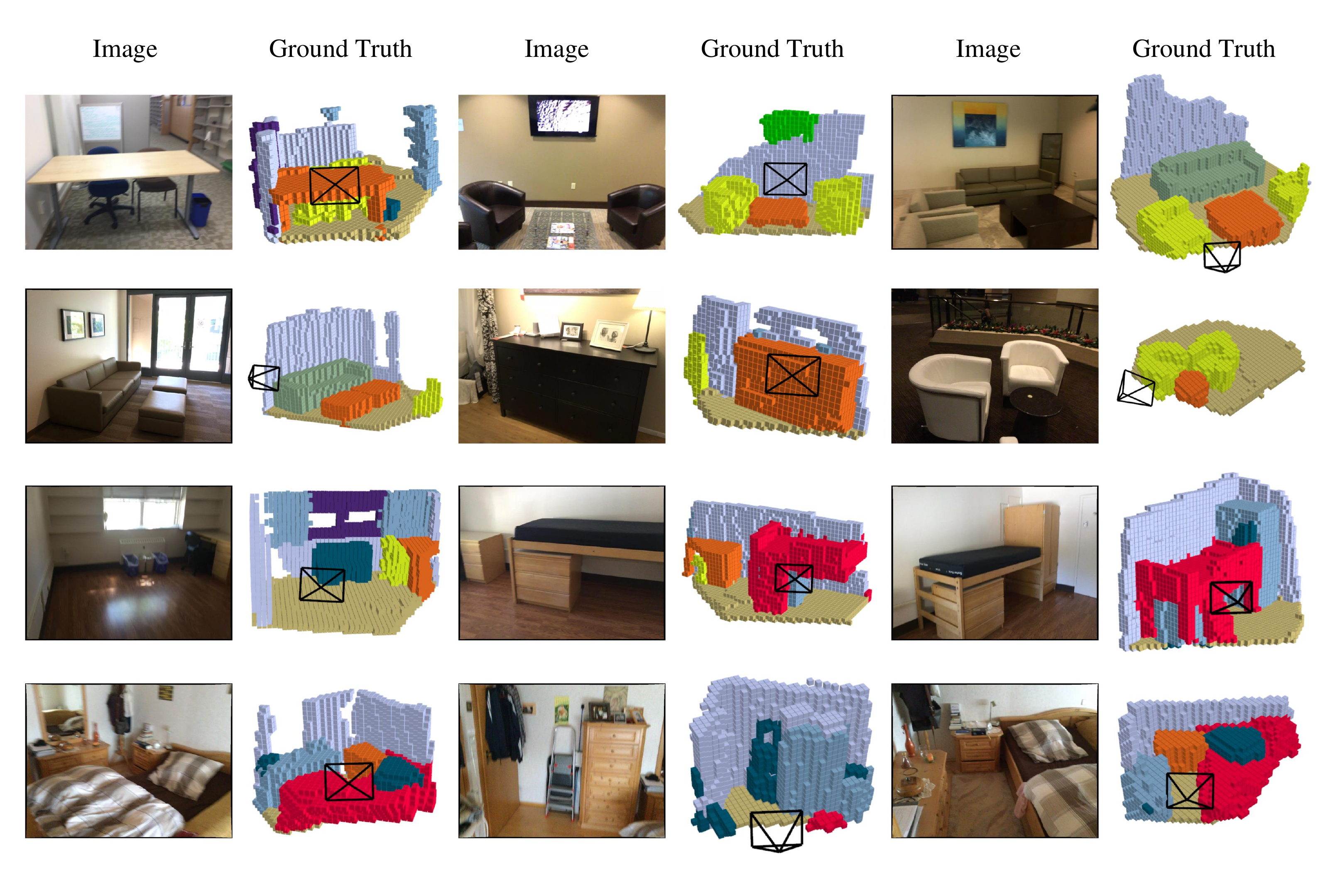}
  \caption{\textbf{Samples Visualization in Occ-ScanNet Benchmark}. The original RGB image is shown in column 1,3 and 5, the corresponding scene voxel labels is shown in column 2, 4 and 6. The first two rows are different views from different scenes and the last two rows each is three different views from the same scene.}
  \label{fig:occscannet_samples}
\end{figure}

\section{Occ-ScanNet Benchmark}

Compared to the previous widely used indoor scene benchmark NYUv2~\cite{silberman2012indoor}, which contains only 795 / 654 for train / test samples, our benchmark boasts 45,755 / 19,764 samples. Our benchmark significantly surpasses NYUv2~\cite{silberman2012indoor} in both data quantity and richness of scene depth, as illustrated in~\cref{fig:dataset_comparison}. This dataset is available at \url{https://huggingface.co/datasets/hongxiaoy/OccScanNet}.

\subsection{Overview}

Occ-ScanNet benchmark features a train/validation split of 45,755 / 19,764  samples. As shown in \cref{fig:occscannet_samples}, Occ-ScanNet exhibits rich diversity in scenes and viewpoints. This diverse dataset not only challenges the task of predicting scene occupancy but also fosters future research endeavors towards developing larger-scale and more versatile occupancy models.

\subsection{Occupancy Label Generation}

\begin{figure}[tb]
  \centering
  \includegraphics[width=\textwidth]{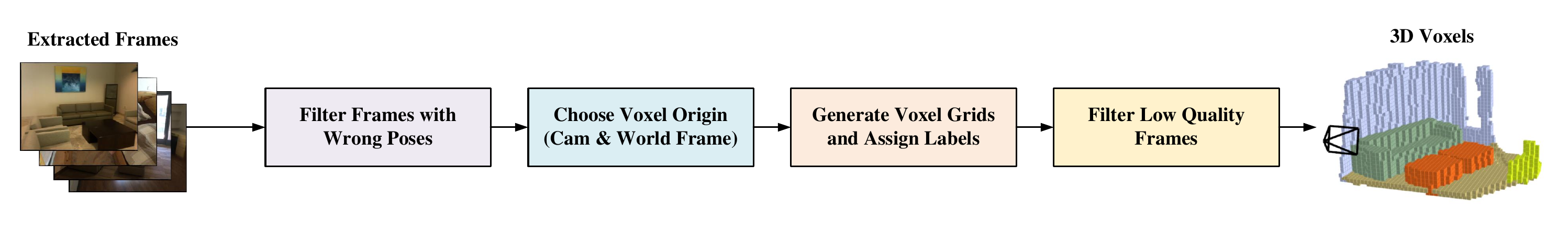}
  \caption{\textbf{Pipeline of Occ-ScanNet dataset label generation}. Color images, depth images, camera intrinsic and poses are extracted from ScanNet scenes. For each scene, 100 frames were sampled and randomly split into training and validation sets with a 7/3 ratio. Frames with invalid camera poses or exceeding scene boundaries were filtered out. Only the area in front of the camera was analyzed, necessitating careful selection of the voxel origin. Voxel were labeled based on their nearest voxel in the CompleteScanNet dataset. Frames with >95\% unknown/empty labels or <2 semantic classes were excluded, resulting in generated 3D voxel labels for each frame.}
  \label{fig:occscannet_pipeline}
\end{figure}

We adhere to the data formulation used in NYUv2~\cite{silberman2012indoor} dataset. The pipeline of occupancy generation process is showed in \cref{fig:occscannet_pipeline}. For generating 3D voxel labels, we initially followed the generation process in CompleteScanNet and then employed manual inspection.

More specifically, from the official ScanNet \cite{dai2017scannet} scenes, we extracted various data components, including color images, depth images, camera intrinsic matrices, and camera poses. A total of 100 frames were sampled from each scene and were randomly divided into training and validation sets with a 7/3 split ratio per scene. 
\subsubsection{Manual inspection.} The manual inspection involves verifying camera poses, camera intrinsics, and voxel positions. We filter out erroneous samples to prevent convergence issues in the model training process.

\subsubsection{Voxel label assignment.} For each frame, only a specific area in front of the camera is defined for analysis. Therefore, the selection of the voxel origin is paramount, as it determines the subsequent coordinates of other voxel. Subsequently, each voxel is assigned a label based on its nearest voxel in the CompleteScanNet dataset.
Additionally, we excluded frames with a ratio of unknown or empty labels exceeding 95\% or frames where the number of semantic label classes was less than two. Consequently, we obtained the generated 3D voxel labels for each frame.


\section{Experiments}

\subsection{Setup}

\subsubsection{Datasets.}

The NYUv2~\cite{silberman2012indoor} dataset provides scenes represented in $240 \times 144 \times 240$ voxels grids, labeled with 13 classes, including 1 for free space, 1 for unknown, and 11 for specific semantics (ceiling, floor, wall, window, chair, bed, sofa, table, tvs, furniture, objects). The dataset consists of 795 / 654 scenes in the train / test splits. The model is trained and evaluated on down-sampled $60 \times 36 \times 60$ voxels.

Occ-ScanNet dataset provides scenes represented in $60 \times 60 \times 36$ voxel grids, labeled with 12 classes including 1 for free space, and 11 for specific semantics (ceiling, floor, wall, window, chair, bed, sofa, table, tvs, furniture, objects). The dataset comprises 45,755 / 19,764 frames in the train / val splits. The model is trained and evaluated on the original scale.

Occ-ScanNet-mini dataset has the same class number and voxel scene size as Occ-ScanNet, except that this mini dataset consists of 4,639 / 2,007 frames in the train / val splits. The model is also trained and evaluated on the original scale.

\subsubsection{Implementation Details.}

We employ a pre-trained EfficientNet-B7~\cite{tan2019efficientnet} as the encoder in our 2D UNet architecture. In the depth branch, we utilize a pre-trained DepthAnything model~\cite{yang2024depth}, which remains frozen during training. Additionally, we integrate a depth loss specific to the depth branch, complementing the other losses outlined in Monoscene~\cite{cao2022monoscene}.
Our model is trained on two datasets: NYUv2~\cite{silberman2012indoor} and Occ-ScanNet. For NYUv2, \methodname{} undergoes 30 epochs of training using AdamW optimizers. Initially, the learning rate is set to 5e-6 for the DepthNet and 1e-4 for other components, with a learning rate reduction by 0.1 at epoch 20. Training on NYUv2 takes approximately 7 hours using 2 NVIDIA L20 GPUs (2 items per GPU).
On the Occ-ScanNet-mini dataset, \methodname{} is trained for 60 epochs under similar learning rate settings, decreasing the rate by 0.1 at epoch 40. This training process takes around 12 hours using 8 L20 GPUs.
For Occ-ScanNet, \methodname{} is trained for 10 epochs with the same learning rate schedule as used for Occ-ScanNet-mini, requiring approximately a day using 8 L20 GPUs.

\subsection{Main Results}

\definecolor{ceiling}{RGB}{214,  38, 40}   %
\definecolor{floor}{RGB}{43, 160, 4}     %
\definecolor{wall}{RGB}{158, 216, 229}  %
\definecolor{window}{RGB}{114, 158, 206}  %
\definecolor{chair}{RGB}{204, 204, 91}   %
\definecolor{bed}{RGB}{255, 186, 119}  %
\definecolor{sofa}{RGB}{147, 102, 188}  %
\definecolor{table}{RGB}{30, 119, 181}   %
\definecolor{tvs}{RGB}{160, 188, 33}   %
\definecolor{furniture}{RGB}{255, 127, 12}  %
\definecolor{objects}{RGB}{196, 175, 214} %

\begin{table}
		\caption{Performance on the Occ-ScanNet dataset}
		\tiny
		\setlength{\tabcolsep}{0.004\linewidth}
		\captionsetup{font=scriptsize}
		\centering
		\begin{tabular}{l|c|c|c c c c c c c c c c c|c}
			\toprule
			Method
			& Input
			& {IoU}
			& \rotatebox{90}{\parbox{2cm}{\textcolor{ceiling}{$\blacksquare$} ceiling}} 
			& \rotatebox{90}{\textcolor{floor}{$\blacksquare$} floor}
			& \rotatebox{90}{\textcolor{wall}{$\blacksquare$} wall} 
			& \rotatebox{90}{\textcolor{window}{$\blacksquare$} window} 
			& \rotatebox{90}{\textcolor{chair}{$\blacksquare$} chair} 
			& \rotatebox{90}{\textcolor{bed}{$\blacksquare$} bed} 
			& \rotatebox{90}{\textcolor{sofa}{$\blacksquare$} sofa} 
			& \rotatebox{90}{\textcolor{table}{$\blacksquare$} table} 
			& \rotatebox{90}{\textcolor{tvs}{$\blacksquare$} tvs} 
			& \rotatebox{90}{\textcolor{furniture}{$\blacksquare$} furniture} 
			& \rotatebox{90}{\textcolor{objects}{$\blacksquare$} objects} 
			& mIoU\\
			\midrule
			MonoScene*~\cite{cao2022monoscene} & $x^{\text{rgb}}$ & 41.60 & 15.17 & \textbf{44.71} & \textbf{22.41} & 12.55 & 26.11 & 27.03 & 35.91 & 28.32 & 6.57 & 32.16 & 19.84 & 24.62 \\
            \methodname{}(Ours) & $x^{\text{rgb}}$ & \textbf{42.16} & \textbf{19.88} & 41.88 & 22.37 & \textbf{16.98} & \textbf{29.09} & \textbf{42.43} & \textbf{42.00} & \textbf{29.60} & \textbf{10.62} & \textbf{36.36} & \textbf{24.61} & \textbf{28.71} \\
			\bottomrule
		\end{tabular}
		\label{tab:perf_scannet}
 \end{table}

\subsubsection{Occ-ScanNet performance.}
\begin{figure}[htb]
  \centering
  \includegraphics[width=0.95\textwidth]{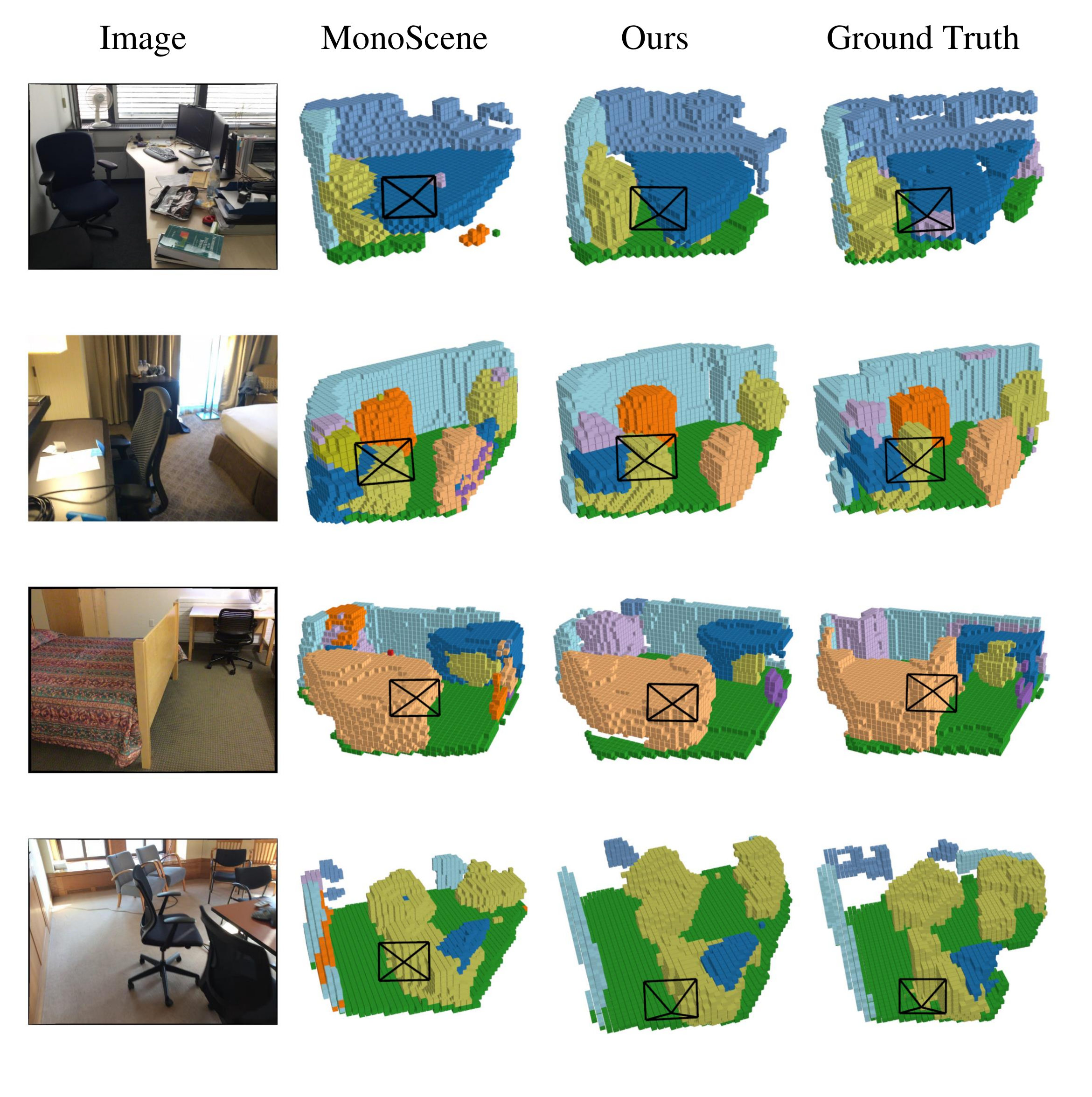}
  \caption{\textbf{Qualitative Analysis on the Occ-ScanNet Dataset.} The input image is displayed on the left, while the predicted scene is shown in the middle two column, and ground truth on the right column.}
  \label{fig:result_show_occscannet}
\end{figure}

We first evaluate our model's performance on the large-scale Occ-ScanNet dataset. As shown in~\cref{tab:perf_scannet}, the results indicate that our method significantly outperforms MonoScene~\cite{cao2022monoscene}. The $*$ denotes results obtained using their code trained on our dataset. In~\cref{fig:result_show_occscannet}, we also conducted qualitative visualization comparisons.

\subsubsection{NYUv2 performance.}
The NYUv2~\cite{silberman2012indoor} dataset serves as a widely used benchmark for indoor scene evaluation. The results in~\cref{tab:perf_nyu} demonstrate that our method achieves state-of-the-art performance.
Our method also demonstrates substantial performance improvement on the NYUv2 dataset, in addition to the gains observed on Occ-ScanNet.
As illustrated in~\cref{fig:result_show}, our \methodname{} demonstrates superior room layout prediction compared to \cite{cao2022monoscene}, particularly evident in rows 1 and 2. This improvement can be attributed to incorporating depth knowledge into the model. Additionally, our method exhibits the capability of learning small objects, such as pictures on the wall in row 3.

\definecolor{ceiling}{RGB}{214,  38, 40}   %
\definecolor{floor}{RGB}{43, 160, 4}     %
\definecolor{wall}{RGB}{158, 216, 229}  %
\definecolor{window}{RGB}{114, 158, 206}  %
\definecolor{chair}{RGB}{204, 204, 91}   %
\definecolor{bed}{RGB}{255, 186, 119}  %
\definecolor{sofa}{RGB}{147, 102, 188}  %
\definecolor{table}{RGB}{30, 119, 181}   %
\definecolor{tvs}{RGB}{160, 188, 33}   %
\definecolor{furniture}{RGB}{255, 127, 12}  %
\definecolor{objects}{RGB}{196, 175, 214} %

\begin{table}
		\caption{Performance on the NYUv2 dataset}
		\tiny
		\setlength{\tabcolsep}{0.003\linewidth}
		\captionsetup{font=scriptsize}
		\centering
		\begin{tabular}{l|c|c|c c c c c c c c c c c|c}
			\toprule
			Method
			& Input
			& {IoU}
			& \rotatebox{90}{\textcolor{ceiling}{$\blacksquare$} ceiling}
			& \rotatebox{90}{\textcolor{floor}{$\blacksquare$} floor}
			& \rotatebox{90}{\textcolor{wall}{$\blacksquare$} wall} 
			& \rotatebox{90}{\textcolor{window}{$\blacksquare$} window} 
			& \rotatebox{90}{\textcolor{chair}{$\blacksquare$} chair} 
			& \rotatebox{90}{\textcolor{bed}{$\blacksquare$} bed} 
			& \rotatebox{90}{\textcolor{sofa}{$\blacksquare$} sofa} 
			& \rotatebox{90}{\textcolor{table}{$\blacksquare$} table} 
			& \rotatebox{90}{\textcolor{tvs}{$\blacksquare$} tvs} 
			& \rotatebox{90}{\textcolor{furniture}{$\blacksquare$} furniture} 
			& \rotatebox{90}{\textcolor{objects}{$\blacksquare$} objects} 
			& mIoU\\
			\midrule
			LMSCNet$^\text{rgb}$~\cite{roldao2020lmscnet} & $\hat{x}^{\text{occ}}$ & 33.93 & 4.49 & 88.41 & 4.63 & 0.25 & 3.94 & 32.03 & 15.44 & 6.57 & 0.02 & 14.51 & 4.39 & 15.88 
			\\
			AICNet$^\text{rgb}$~\cite{li2020anisotropic} & $x^{\text{rgb}}$, $\hat{x}^{\text{depth}}$ & 30.03 & 7.58 & 82.97 & 9.15 & 0.05 & 6.93 & 35.87 & 22.92 & 11.11 & 0.71 & 15.90 & 6.45 & 18.15 
			\\
			3DSketch$^\text{rgb}$~\cite{chen20203d}& $x^{\text{rgb}}$, $\hat{x}^{\text{TSDF}}$  & 38.64 & 8.53 & 90.45 &	9.94 & 5.67 & 10.64 & 42.29 & 29.21& 13.88 & 9.38 & 23.83 & 8.19 & 22.91
			\\
			MonoScene~\cite{cao2022monoscene} & $x^{\text{rgb}}$ & 42.51 & 8.89 & 93.50 & 12.06 & 12.57 & 13.72 & 48.19 & 36.11 & 15.13 & 15.22 & 27.96 & 12.94 & 26.94\\
                NDC-Scene~\cite{yao2023ndc} & $x^{\text{rgb}}$ & 44.17 & 12.02 & \textbf{93.51} & 13.11 & 13.77 & 15.83 & 49.57 & 39.87 & 17.17 & 24.57 & 31.00 & 14.96 & 29.03 \\
            \methodname{}(Ours) & $x^{\text{rgb}}$ & \textbf{47.11}& \textbf{14.21} & 93.47 & \textbf{15.89} & \textbf{15.14} & \textbf{18.35} & \textbf{50.01} & \textbf{40.82} & \textbf{18.25} & \textbf{25.90} & \textbf{34.08} & \textbf{17.67}& \textbf{31.25} \\
			\bottomrule
		\end{tabular}

		\label{tab:perf_nyu}
 \end{table}

\begin{figure}[tb]
  \centering
  \includegraphics[width=\textwidth]{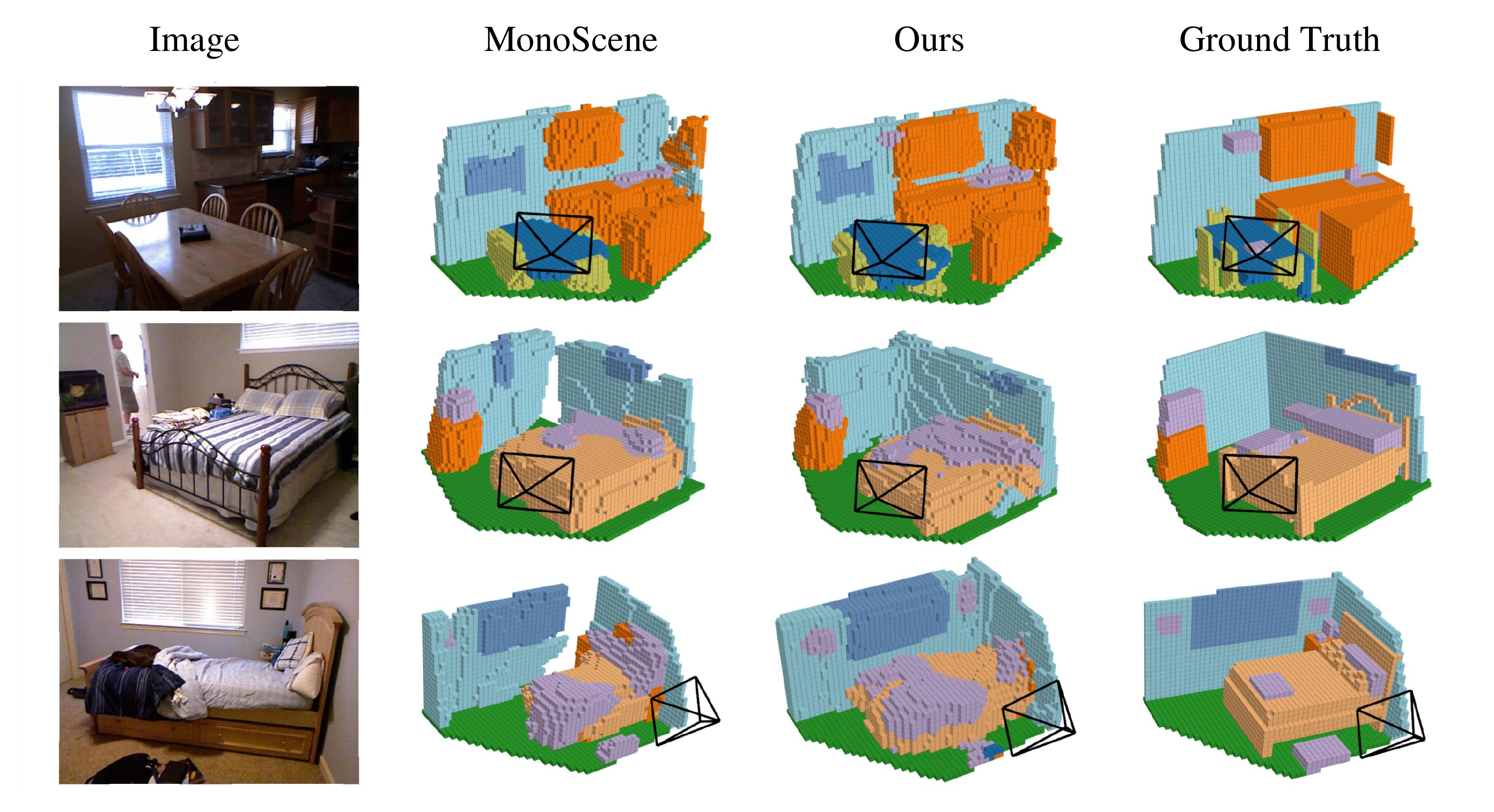}
  \caption{\textbf{Qualitative Analysis on the NYUv2 Dataset}: The input is displayed on the left, with the camera viewing frustum shown in every image. \methodname{} adeptly captures the scene layout and delineates various semantic instances more accurately. Notably, it excels in reconstructing the corners of walls (row 2) and accurately representing small objects like pictures on the wall (row 3).}
  \label{fig:result_show}
\end{figure}

\subsection{Ablation Study}

We conduct ablation studies on the NYUv2~\cite{silberman2012indoor} and Occ-ScanNet-mini datasets to validate the effectiveness of our model design.

\subsubsection{Depth fusion.} 
Inspired by BEVDepth~\cite{li2023bevdepth}, we initially adopt the integration of BEV depth information, denoted as `+bev-depth'. However, we observed that while BEV depth works well for 3D detection in outdoor scenes, directly applying it to indoor scenes is less effective. We speculate that this discrepancy arises because height information is more crucial in indoor scenes, where indoor heights often correspond to different 3D structures and semantics. Therefore, we adopt voxel depth fusion, denoted as `+voxel-depth'. Specifically, we project each 3D voxel centroid to a frustum grid to sample the depth distribution probability of the voxel. We then weight the 2D feature at that pixel coordinate as the voxel feature. The results in~\cref{tab:ablation_nyu_occ_scannet} further validate that the voxel depth fusion approach is more effective.

The BEV method usually performs better in detecting ceilings and floors but performs poorly in detecting furniture such as windows, walls, televisions, beds, sofas, and tables. Ceilings and floors usually have relatively flat and regular geometric shapes, which makes their projections clearer and more consistent from a BEV perspective. In contrast, objects such as windows, walls, and televisions typically have more complex geometric shapes. In the BEV perspective, the projections of these objects may become blurry or overlapping, making it difficult to accurately distinguish and detect.

\begin{table}
		\caption{Ablation study of depth fusion on the NYUv2 and Occ-ScanNet-mini dataset}
		\setlength{\tabcolsep}{0.01\linewidth}
		\captionsetup{font=scriptsize}
		\centering
		\begin{tabular}{l|c|c c|c c}
			\toprule
            \multirow{2}{*}{Method} & \multirow{2}{*}{Input} & \multicolumn{2}{c|}{NYUv2} & \multicolumn{2}{c}{Occ-ScanNet-mini} \\
			&
			& IoU
			& mIoU 
            & IoU
            & mIoU \\
			\midrule
			baseline & $x^{\text{rgb}}$ & 42.27 & 27.13 & 50.94 & 38.95 \\
                + bev-depth & $x^{\text{rgb}}$ & 42.67 & 27.14 & \textbf{51.58} & 38.48 \\
                + voxel-depth & $x^{\text{rgb}}$ & \textbf{47.11} & \textbf{31.25} & 51.03 & \textbf{39.08} \\
			\bottomrule
		\end{tabular}

		\label{tab:ablation_nyu_occ_scannet}
 \end{table}

\subsubsection{Different depth pre-trained models.} 

\begin{table}
		\caption{Ablation of depth information on the NYUv2 dataset}
		\scriptsize
		\setlength{\tabcolsep}{0.01\linewidth}
		\captionsetup{font=scriptsize}
		\centering
        \begin{tabular}{ccccc}
            \toprule
            depth-method & learned & multi-scale & IoU & mIoU \\
            \midrule
            GT & & \checkmark & 53.98 & 34.47 \\  \midrule
            
            ZoeDepth & \checkmark &  & 45.18 & 29.15 \\
            ZoeDepth & \checkmark & \checkmark & 45.24 & 29.40 \\  \midrule

            DepthAnything & &  & 44.77 & 29.28 \\
            DepthAnything & & \checkmark & 45.48 & 29.57 \\

            DepthAnything & \checkmark & \checkmark & 46.94 & 31.02 \\
            DepthAnything & \checkmark &  & \textbf{47.11} & \textbf{31.25} \\

            \bottomrule
        \end{tabular}
		\label{tab:ablation2_occ_mini}
 \end{table}

We conducted ablation experiments on the depth method using the Occ-ScanNet-mini dataset. Initially, we observed the upper bound of model performance when ground truth (GT) depth information was provided. Subsequently, we initialized our depth module using ZoeDepth~\cite{bhat2023zoedepth} and Depth-Anything~\cite{yang2024depth}. Experimental results indicate that utilizing Depth-Anything~\cite{yang2024depth} yields better performance. Moreover, our approach allows for further enhancement by fine-tuning a pre-trained depth module, resulting in improved performance.

\subsubsection{Multi-scale depth.} 
In ~\cref{tab:ablation2_occ_mini}, we investigate the impact of using multi-scale depth distribution on the NYUv2~\cite{silberman2012indoor} dataset.
Specifically, the multi-scale depth distribution is used to weight the multi-scale 3D voxel feature. Meanwhile, we tried a method that only uses the single-scale depth distribution map to generate the 3D voxel depth matrix. The used depth distribution map has a scale of $\frac{1}{8}$ original image scale. Only this single-scale 3D voxel matrix is used to weigh the summed-up 3D voxel feature out of the FLoSP module. Without multi-scale depth, all 3D features are simply added up and a single depth distribution is used for weighting. This method ignores the different importance that features at different scales may have, as all features at all scales are treated equally and all depth information is weighted by a unified depth distribution. With multi-scale depth, the 3D features of each scale are first weighted with the depth distribution of their corresponding scale, and then these weighted features are added up. This method takes into account the differences in the importance of features at different scales, as each scale's feature is weighted based on its corresponding depth distribution.

In real-world scenarios, various objects and structures (such as ceilings, walls, furniture, etc.) often exhibit different scales. Hence, accounting for scale variations is crucial for accurately modeling these objects. Multi-scale depth can effectively capture these scale changes by assigning different weights to features at each scale. Depth information is indispensable for 3D scene completion. Leveraging the depth distribution information at each scale allows us to weight the corresponding 3D features, enabling the model to better understand and represent the objects in the scene and their relationships.

\subsection{Discussion}

\subsubsection{Data scaling up.} 
In ~\cref{tab:data_scale}, we delve into the impact of data volume on our method's performance by scaling up the number of scene samples.
We trained the model on Occ-ScanNet using 10\% and 100\% scene samples respectively, and tested the model performance. The results indicate that larger dataset exhibits more significant gains.

 \begin{table}
		\caption{Scaling up comparison on the Occ-ScanNet}
        \small
		\setlength{\tabcolsep}{0.008\linewidth}
		\captionsetup{font=scriptsize}
		\centering
		\begin{tabular}{c|c| c c}
			\toprule
                 \multirow{2}{*}{Data} & \multirow{2}{*}{Depth Branch}& \multicolumn{2}{c}{Ours}  \\ 
			& & IoU
			& mIoU  \\
			\midrule
 			 10\% & \checkmark & 21.79 & 9.57 \\
             100\% & \checkmark & 42.16 & 28.71 \\
			\bottomrule
		\end{tabular}
		\label{tab:data_scale}
 \end{table}

\subsubsection{Limitations and future works.}
Although our model can estimate 3D occupancy effectively with the assistance of depth information, semantic learning still faces challenges due to class imbalance issues. Furthermore, our proposed Occ-ScanNet only considers 11 common semantic classes, which may not fully capture the diversity of categories present in real-world scenarios. 
Despite the significant increase in data volume compared to previous datasets, Occ-ScanNet remains limited in the number of scenes it covers.
In future work, we will focus more on semantic exploration and conduct tests in more generalized scenarios.

\section{Conclusion}
In this paper, we introduce \methodname{}, a novel method for monocular 3D occupancy prediction. \methodname{} utilizes a D-FLoSP (Dual Feature Line of Sight Projection) and a multi-scale feature fusion strategy to address variations in scene and object sizes, specifically tailored for indoor scenes.
Additionally, we present a new benchmark, Occ-ScanNet, aimed at fostering scalable studies of indoor scenes. We hope that this dataset will attract more attention to research on indoor occupancy prediction.
While achieving promising results in 3D structure prediction, accurately identifying semantics remains a significant challenge in the future.

\section*{Acknowledgement}
This work was supported in part by the National Key R\&D Program of China (No. 2022ZD0160102), the National Natural Science Foundation of China (No. U21B2042, No. 62320106010), and in part by the 2035 Innovation Program of CAS, and the InnoHK program.

\bibliographystyle{splncs04}

\end{document}